\documentclass[letterpaper, 10 pt, conference]{ieeeconf}  

\IEEEoverridecommandlockouts                              

\usepackage[backend=biber,
            hyperref=false,
            url=false,
            isbn=false,
            doi=false,
            backref=false,
            style=numeric-comp,
            sortcites,
            block=none]{biblatex}

\usepackage[english]{babel}
\usepackage[T1]{fontenc}

\usepackage[nolist]{acronym}
\usepackage{amsmath}
\usepackage[font={small}]{caption}
\usepackage{subcaption}
\usepackage[export]{adjustbox}
\usepackage{amssymb}
\usepackage{bm}
\usepackage{booktabs}
\usepackage{booktabs}
\usepackage{balance}
\usepackage{csquotes}
\usepackage{color}
\usepackage{colortbl}
\usepackage{dsfont}
\usepackage{esvect}
\usepackage{float}
\usepackage{graphicx}
\usepackage{hyperref}
\usepackage{import}
\usepackage{mathtools}
\usepackage{multirow}
\usepackage{multicol}
\usepackage{multirow}
\usepackage{flushend}
\usepackage{outline}
\usepackage{paralist}
\usepackage{pifont}
\usepackage{siunitx}
\usepackage{stmaryrd}
\usepackage{systeme}
\usepackage{soul}
\usepackage{optidef}
\usepackage{url}
\usepackage{tikz}
\usepackage{tabularx}
\usepackage{verbatim}
\usetikzlibrary{tikzmark}

\usepackage[normalem]{ulem}

\newcommand{\listingsttfamily}{\fontfamily{pcr}\small}
\usepackage{listings}
\usepackage{algorithmic}

\usepackage{graphicx}
\usepackage{amsmath, xparse}
\usepackage{siunitx}
\usepackage{bbding}
\usepackage{makecell}
\usepackage{tabu}
\usepackage{mathtools}

\usepackage{listings}
\usepackage{algorithmic}
\usepackage{algorithm}
\usepackage{multirow}

\usepackage{adjustbox}
\usepackage{makecell}
\usepackage{textcomp}
\usepackage{xcolor}         
\usepackage{pifont}
\definecolor{darkgreen}{RGB}{0,110,0}
\definecolor{darkred}{RGB}{170,0,0}
\definecolor{codeblue}{rgb}{0.25,0.5,0.5}
\definecolor{keyword}{rgb}{0.8, 0.25, 0.5}
\def\greencheckmark{\textcolor{darkgreen}{\checkmark}}

\newcommand{\note}[1]{\textcolor{black}{#1}}
\newcommand{\noteb}[1]{\textcolor{black}{#1}}

\title{\LARGE \bf ZeroBP: Learning Position-Aware Correspondence\\for Zero-shot 6D Pose Estimation in Bin-Picking} 

\author{
\authorblockA{Jianqiu Chen\textsuperscript{*}, Zikun Zhou\textsuperscript{*}, Xin Li, Ye Zheng, Tianpeng Bao and Zhenyu He\textsuperscript{†}}
\thanks{
    Jianqiu Chen and Zhenyu He (Corresponding author $\dagger$) are with Harbin Institute of Technology, Shenzhen, China (e-mail: zhenyuhe@hit.edu.cn). Zikun Zhou and Xin Li are with Pengcheng Laboratory, Shenzhen, China. Ye Zheng is with JD.com, Inc. Tianpeng Bao is with SenseTime Research. 
    \textsuperscript{*}Jianqiu Chen and Zikun Zhou contributed equally.
    }
}

\begin{document}

\maketitle
\thispagestyle{empty}
\pagestyle{empty}


\begin{abstract}
Bin-picking is a practical and challenging robotic manipulation task, where accurate 6D pose estimation plays a pivotal role. The workpieces in bin-picking are typically textureless and randomly stacked in a bin, which poses a significant challenge to 6D pose estimation. Existing solutions are typically learning-based methods, which require object-specific training. Their efficiency of practical deployment for novel workpieces is highly limited by data collection and model retraining. Zero-shot 6D pose estimation is a potential approach to address the issue of deployment efficiency. Nevertheless, existing zero-shot 6D pose estimation methods are designed to leverage feature matching to establish point-to-point correspondences for pose estimation, which is less effective for workpieces with textureless appearances and ambiguous local regions. In this paper, we propose ZeroBP, a zero-shot pose estimation framework designed specifically for the bin-picking task. ZeroBP learns Position-Aware Correspondence (PAC) between the scene instance and its CAD model, leveraging both local features and global positions to resolve the mismatch issue caused by ambiguous regions with similar shapes and appearances. Extensive experiments on the ROBI dataset demonstrate that ZeroBP outperforms state-of-the-art zero-shot pose estimation methods, achieving an improvement of 9.1\% in average recall of correct poses.
\end{abstract}
\section{Introduction}
\label{sec:intro}
6D object pose estimation plays a pivotal role for robots to interact with objects in real-world environments robustly~\cite{MVBPICRA, ST6DICRA, st6deccv, robi}. As a crucial yet challenging practice application, bin-picking requires robots to first accurately detect the workpieces and estimate their 6D poses before grasping. Herein the workpieces are typically textureless and stacked in a bin randomly. Numerous learning-based methods~\cite{mpaae, aae, ST6DICRA, st6deccv,miretr, dcnet} have been proposed to \noteb{handle this task} by training an object-specific model \noteb{for each workpiece}. When deployed for a novel object, these methods need days to collect real-world data or generate synthetic data for this object and retrain the model. Such an expensive and time-consuming deployment process limits the practical application of these approaches. 
\note{Besides, these methods may fail when target objects are only available during inference.}

\begin{figure}[t]
    \centering
    \includegraphics[width=0.49 \textwidth]{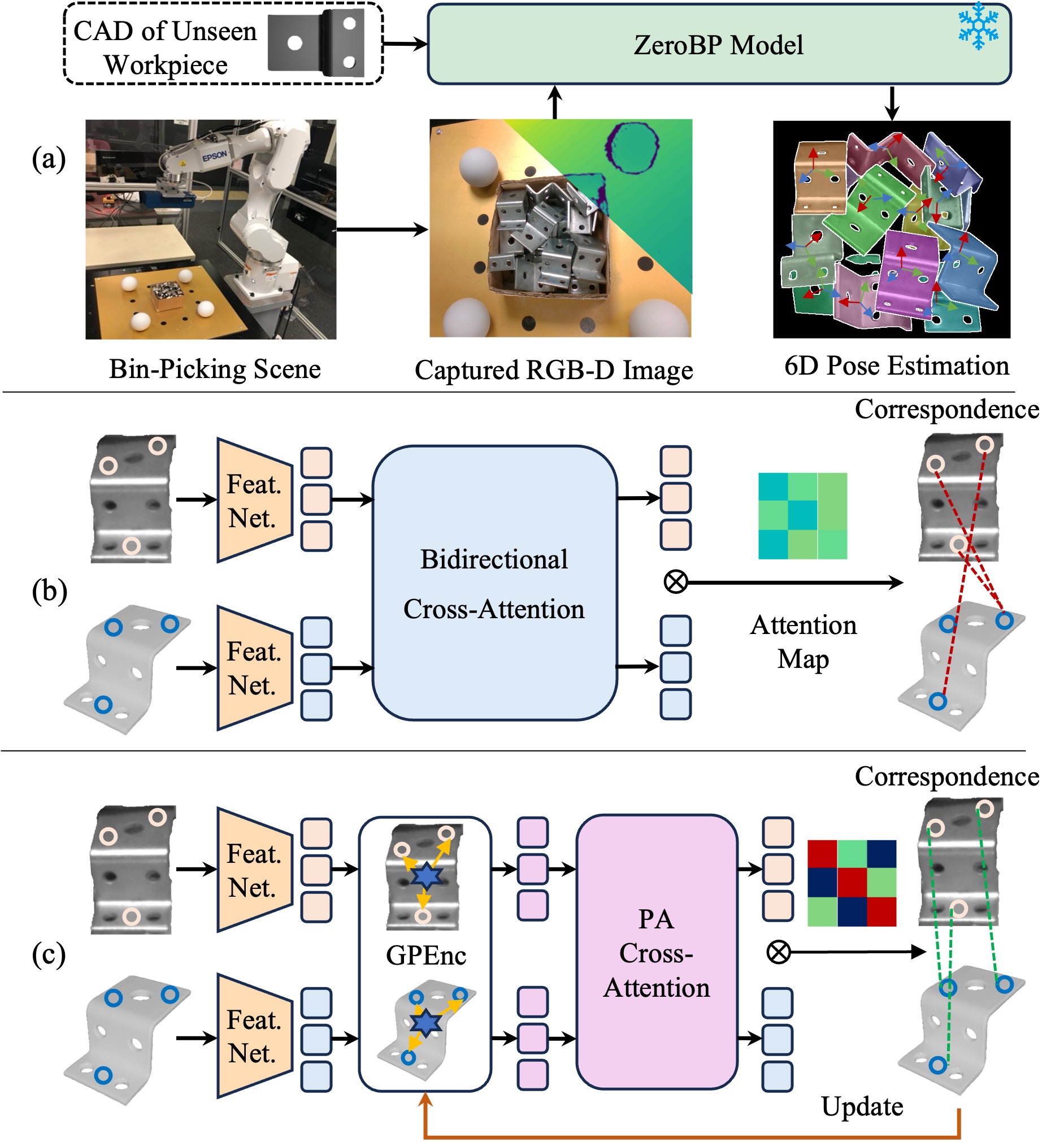}
    \caption{(a) The brief overview of ZeroBP. Given the CAD model of an unseen object, the model can be generalized to the real-world bin-picking scene for 6D pose estimation without requiring retraining. (b) The existing local feature matching for 6D pose estimation. (c) The proposed learning position-aware correspondence for 6D pose estimation. \note{"GPEnc" refers to Globally Positional Encoding, and "PA Cross-Attention" refers to Position-Aware Cross-Attention.}}
    \label{fig:fig1}
    \vspace{-2mm}
\end{figure}

Recently, zero-shot 6D pose estimation~\cite{chen2023zeropose, sam6d, labbe2022megapose} has been widely discussed. It aims to detect unseen objects and estimate their pose given only the CAD models, alleviating the deployment efficiency issue. Nevertheless, existing zero-shot methods are typically designed for daily applications, such as augmented reality. They mainly leverage the rich texture information on daily objects and perform RGB-D feature matching between the detected instance and the CAD model for pose estimation. However, for the manufactured workpieces in bin-picking, which lack sufficient texture and contain ambiguous local regions with similar shapes and appearances, these zero-shot methods driven by local feature matching usually suffer from mismatches, as shown in Fig.~\ref{fig:fig1} (b). 
\noteb{The pose estimation accuracy of these zero-shot methods can be substantially degenerated by the noisy correspondence.} It is necessary to design a zero-shot method specifically for bin-picking.

In this paper, we propose a zero-shot 6D pose estimation method for bin-picking, named ZeroBP, which learns robust Position-Aware Correspondence (PAC) between the scene instance and CAD model, as shown in Fig.~\ref{fig:fig1} (a) and (c). It leverages both the local features and global positions to distinguish between two points with similar shapes and appearances but located far apart. The key to PAC is encoding the global position of the heterogeneous point clouds, \emph{i.e.}, the points of the scene instance and CAD model, in a comparable manner. That is, we need to project the heterogeneous point clouds into a shared coordinate system to encode the global position, whereas the projection itself requires knowing the pose. This raises an intriguing cyclical dependency issue between pose and global position. To address this issue, we start with an approximate initial pose and alternately refine the pose and global position iteratively. \noteb{To represent the point position, we propose a multiplicative positional encoding, defined as the directional vector from the object centroid to the surface point. To leverage the point position, we design a position-aware cross-attention, which effectively integrates positional encodings with local features for correspondence modeling.} During alternate refinement, the multiplicative positional encoding and position-aware cross-attention allow ZeroBP to gradually establish robust correspondence.

\noteb{Experiments on the real-world dataset ROBI~\cite{robi} demonstrate that ZeroBP
outperforms state-of-the-art zero-shot 6D pose estimation methods, achieving an improvement of 9.1\% in average recall of correct pose.} To summarize, we make the following contributions:
\begin{itemize}
    \item \note{We propose a zero-shot 6D pose estimation method for bin-picking, which learns robust position-aware correspondence to alleviate mismatch in ambiguous regions.}
    \item We design the multiplicative positional encoding and position-aware cross-attention to effectively represent and leverage the global position for pose estimation.
    \item Extensive experiments show that our method substantially improves the pose estimation accuracy and outperforms state-of-the-art zero-shot methods.
\end{itemize}

\begin{figure*}[!t]
    \centering
    \includegraphics[width=0.975 \textwidth]{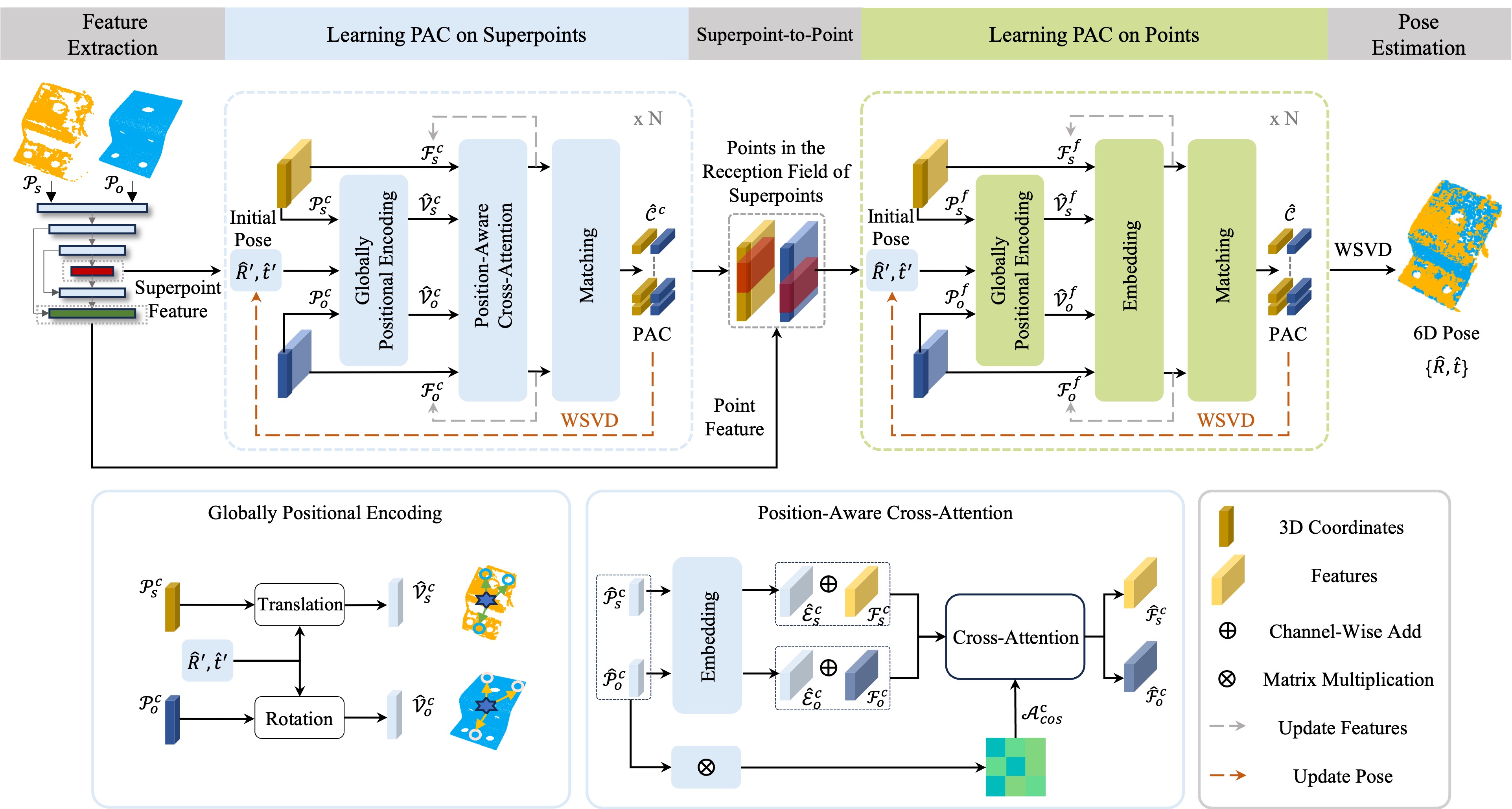}
    \caption{Overview of learning Position-Aware Correspondence (PAC) for zero-shot 6D pose estimation in bin-picking. \noteb{We introduce the globally positional encoding and position-aware cross-attention modules to learn robust PAC, alleviating the mismatch issue in textureless bin-picking workpieces. Overall, we adopt a coarse-to-fine strategy to establish the correspondence from superpoints to points.}}
    \label{fig:overall_framwork}
    \vspace{-2mm}
\end{figure*}

\vspace{-1mm}
\section{Related Work}
\label{sec:related work}
\subsection{Object-specific 6D pose estimation models in bin-picking}
Compared with the general household object pose estimation, the bin-picking objects are often manufactured workpieces, which lack sufficient texture and stacking in a bin.
To estimate their poses, existing methods~\cite{kleeberger2020single, aae,dcnet,mpaae,miretr, MVBPICRA, uni6dv2, chen2023geo6d, pprnet} usually prepare training data for training an object-specific model for pose estimation.
However, these training data are expensive to prepare and annotate 6D poses.
To solve that, current methods~\cite{st6deccv, ST6DICRA} introduce a domain adaptation strategy, named sim2real. This sim2real strategy enables the model training on the synthetic images of target objects and uses the real-world unannotated images for fine-tuning in a self-supervised manner. However, although it reduces the deployment cost in data preparation, the model still has an expensive deployment in training the model when an unseen object coming.
Such an expensive and time-consuming deployment process limits the practical application of these approaches.
\subsection{Zero-shot 6D pose estimation models}
Recent works~\cite{chen2023zeropose, sam6d, nguyen2024gigapose, ornek2023foundpose, FoundationPose, shugurov2022cvpr:osop, gcpose, ausserlechner2023zs6d, park2020latentfusion} demonstrate the strong generalization capability for zero-shot 6D pose estimation in household objects and scenes.
After the pertaining on the seen objects, the zero-shot model can generalize to novel unseen objects without retraining the model.
These methods~\cite{chen2023zeropose,sam6d, nguyen2024gigapose, ornek2023foundpose, shugurov2022cvpr:osop} estimate the pose by matching the visual features in objects from the scene image and the template images rendered from the CAD model. 
For the manufactured target objects in bin-picking, which lack sufficient texture and contain ambiguous local regions with the same structure and appearance, these zero-shot methods usually suffer from mismatching and fail to estimate the pose accurately. Therefore, existing zero-shot methods still cannot be effectively applied to the bin-picking task.

\subsection{Positional encoding}
Positional encoding is a crucial component in feature extraction and interaction networks~\cite{vaswani2017attention, liu2022petr, pointnet, pointnet++, qin2023geotransformer}. It embeds positional information into features, allowing for feature attention with positional constraints. However, the existing positional encoding strategies are typically designed for the homogeneous source input. When processing heterogeneous source feature attention, they primarily rely on feature similarity without incorporating positional encoding. 
This makes the attention mechanism unreliable in features with insufficient discriminative, ultimately leading to a failure in estimating correct correspondence through feature matching.
\section{Method}
\subsection{Preliminaries}
\noindent\textbf{Problem formulation.}
Zero-shot 6D pose estimation in bin-picking aims to detect the target \note{objects} in the input RGB-D image $I\! \in\! \mathbb{R}^{H \times W \times 4}$ of the scene, based on the given CAD model $\mathcal{O}$, and to predict their relative 6D pose transformations $R \in SO(3)$ and $t \in \mathbb{R}^3$ \emph{w.r.t} the CAD model.

\vspace{1mm}
\noindent\textbf{Zero-shot 6D pose estimation pipeline.} Given that directly estimating the pose of randomly stacked \note{workpieces} is quite challenging, we adopt the popular two-stage pipeline~\cite{chen2023zeropose, sam6d, FoundationPose, nguyen2024gigapose}. It simplifies pose estimation into two sub-tasks: 1) object detection and 2) point registration via RGB-D feature matching. Next, we briefly describe this pipeline.

The pipeline usually employs a CAD-prompt segmentation model~\cite{chen2023zeropose, sam6d} to detect each instance of the target object from $I$, obtaining the 2D mask of each instance. With the mask and depth map, it extracts the point cloud $\mathcal{P}_s = \{p_i \in \mathbb{R}^3\}_{i=1}^{N_s}$ for each instance of the target object in the camera coordinate system. Herein $N_s$ is the valid pixels in the mask. 

With the point cloud of each instance, the pose estimation is formulated as a point registration problem. Defining an object coordinate system in CAD and denoting the point cloud uniformly sampled from CAD as $\mathcal{P}_o = \{q_i \in \mathbb{R}^3\}_{i=1}^{N_o}$, where $N_o$ is the sampled point number. The pipeline first estimates the point-to-point correspondence $\mathcal{C}$ between $\mathcal{P}_s$ and $\mathcal{P}_o$ through \noteb{feature matching of the color point clouds}. Then, it calculates the pose parameter $R$ and $t$ by solving: 
\begin{equation}
\label{equ:solve_pose}
\min_{R, t} \sum\nolimits_{({p}_{x_i}, {q}_{y_i}) \in \mathcal{C}} \lVert R {p}_{x_i} + t - {q}_{y_i} \rVert^2_2.
\end{equation}

The \noteb{color point feature matching method} highly relies on discriminative textures to model the correspondence. It struggles to deal with the workpieces lacking sufficient texture and containing ambiguous local regions with similar shapes and appearances.

\subsection{Overview of ZeroBP}
\noteb{We first provide an overview of ZeroBP, a zero-shot 6D pose estimation method customized for bin-picking. It performs zero-shot 6D pose estimation via two steps}: 1) picking-box-aware workpiece detection, and 2) robust point registration with position-aware correspondence.

\vspace{1mm}
\noindent\textbf{Picking-box-aware workpiece detection.}
ZeroBP exploits the prior information of the bin-picking task to improve the detection accuracy. Specifically, it employs a zero-shot detector~\cite{groundingdino} with the tailored text description prompt to pre-locate the picking box, and then detect the workpieces within the picking box following~\cite{chen2023zeropose}, avoiding the interference on workpiece detection from cluttered backgrounds.

\vspace{1mm}
\noindent\textbf{Point registration with position-aware correspondence.} With the point clouds of the instance and CAD model, $\mathcal{P}_s$ and $\mathcal{P}_o$, ZeroBP calculates the correspondence $\mathcal{C}$ between them and accordingly solve Eq.~\eqref{equ:solve_pose} to obtain $R$ and $t$. 

To address the challenge posed by workpieces with insufficient texture and ambiguous local regions, we propose to learn Position-Aware Correspondence (PAC) based on both local features and global positions of the points, as shown in Figure~\ref{fig:overall_framwork}. \noteb{The key to PAC is encoding the global position of the heterogeneous point clouds, $\mathcal{P}_s$ and $\mathcal{P}_o$ in a comparable manner so that they can be used to guide the model to discriminate the local regions with similar shapes and appearances. Comparability means that we first need to project $\mathcal{P}_s$ and $\mathcal{P}_o$ to a shared coordinate system, which however requires knowing the relative pose between $\mathcal{P}_s$ and $\mathcal{P}_o$.} This raises an intriguing cyclical dependency issue: accurate global position is necessary for robust pose estimation, yet acquiring the global position itself depends on the pose. \noteb{To handle this problem, we introduce an initial pose and alternately refine the global position and pose step-by-step. During step-wise refining, we use position-aware cross-attention to fully integrate the global positions and local features to establish robust correspondence.}

We adopt a coarse-to-fine correspondence modeling strategy, following~\cite{qin2023geotransformer}. As shown in Fig.~\ref{fig:overall_framwork}, we first use KPConv-FPN~\cite{thomas2019kpconv} to extract the hierarchical features of $\mathcal{P}_s$ and $\mathcal{P}_o$, obtaining the coarse-level superpoints $\mathcal{P}^c_s$, $\mathcal{P}^c_o$ and their features $\mathcal{F}^c_s \in \mathbb{R}^{\lvert {\mathcal{P}^c_s} \vert \times d^c}$, $\mathcal{F}^c_o \in \mathbb{R}^{\lvert {\mathcal{P}^c_o} \vert \times d^c}$ and the fine-level points $\mathcal{P}^f_s$, $\mathcal{P}^f_o$ and their features $\mathcal{F}^f_s \in \mathbb{R}^{\lvert {\mathcal{P}^f_s} \vert \times d^f}$, $\mathcal{F}^f_o \in \mathbb{R}^{\lvert \mathcal{P}^f_o \vert \times d^f}$, where $d^c$ and $d^f$ are feature dimensions. 
\note{Note that we technically feed color point clouds to KPConv-FPN.}
\noteb{With the features, ZeroBP first models the coarse-level position-aware correspondence on representative superpoints, and then accordingly models the fine-level position-aware correspondence on the dense points. Subsequently, it solves Eq.~\ref{equ:solve_pose} to obtain the final predicted pose parameter $\hat{R}$ and $\hat{t}$.}


\subsection{Global positional encoding of heterogeneous point clouds}
The global position aims to aid in distinguishing between two points having similar shape and appearance but located far apart. Different from existing positional encoding~\cite{transformer,qin2023geotransformer} used for homogeneous sources, our positional encoding is designed for processing heterogeneous point clouds, \emph{i.e.}, $\mathcal{P}_s$ and $\mathcal{P}_o$. To address the above-mentioned cyclical dependency between the pose and global position, we first estimate an approximate initial pose and then alternately refine the global position and pose step-by-step. For the positional encoding itself, we propose a multiplicative positional encoding, defined as the directional vector from the object centroid to the surface point. It can be seamlessly integrated into correspondence modeling.

\vspace{1mm}
\noindent\textbf{Initial pose estimation.}
To obtain an initial pose, we estimate the pose of the target workpieces based on only the local features. Specifically, we perform feature matching between the coarse-level features of the scene instance $\mathcal{F}_s^c$ and the CAD model $\mathcal{F}_o^c$ by calculating their cosine similarity matrix. Then, we select the superpoint pairs with top $K$ similarities to obtain the correspondence $\hat{\mathcal{C}}^c$. 
With the correspondence $\hat{\mathcal{C}}^c$, we minimize Eq.~\eqref{equ:solve_pose} to generate the initial pose $\hat{R}'$, $\hat{t}'$ by the weighted singular value decomposition (WSVD)~\cite{svd} algorithm.

\vspace{1mm}
\noindent\textbf{Position encodings via directional vector.}
With the initial pose, we separately transform the heterogeneous superpoints, $\mathcal{P}^{c}_s$ and $\mathcal{P}^{c}_o$, to obtain the directional vectors within a shared coordinate system. Without loss of generality, we set the above-mentioned centroid defining the direction as the origin of the object coordinate system in the CAD model.
In this way, we translate the superpoints of the scene instance $\mathcal{P}_s^c$ by $\hat{t}'$ and rotate the superpoints of the CAD model $\mathcal{P}_o^c$ for alignment. Then we can obtain the positional encodings $\hat{\mathcal{V}}_s^c \in \mathbb{R}^{\lvert {\mathcal{P}^c_s}\rvert \times 3}$ and $\hat{\mathcal{V}}_o^c \in \mathbb{R}^{\lvert {\mathcal{P}^c_o}\rvert \times 3}$ after normalization, \emph{i.e.,} the directional vectors of the superpoints:
\note{\begin{equation}
\label{eq:coe}
    \hat{\mathcal{V}}_s^c = \frac{\mathcal{P}_s^c - \hat{t}'}{\lVert \mathcal{P}_s^c - \hat{t}'\rVert^2_2}, \hat{\mathcal{V}}_o^c = \frac{\hat{R}' \mathcal{P}_o^c}{\lVert \hat{R}'  \mathcal{P}_o^c\rVert^2_2}.              
\end{equation}}
Eq.~\eqref{eq:coe} formulates the positional encodings for superpoints, and the positional encoding for dense points can be done in the same way. Fig.~\ref{fig:overall_framwork} depicts the positional encoding process.

The directional vectors are multiplicative positional encodings. The inner product between two vectors represents the angle between the corresponding directions, measuring the position consistency between the superpoints from $\mathcal{P}_{s}^{c}$ and $\mathcal{P}_{o}^{c}$. This multiplicative nature means that the positional encodings can be directly integrated into the cross-attention between the features of $\mathcal{P}^{c}_s$ and $\mathcal{P}^{c}_o$, allowing learning correspondence based on both local features and global positions.

\vspace{1mm}
\noindent\textbf{Alternate refinement between pose and global position.} With the above positional encoding, ZeroBP models the position-aware correspondence between $\mathcal{P}_{s}^{c}$ and $\mathcal{P}_{o}^{c}$ and uses WSVD to solve Eq.~\eqref{equ:solve_pose} to obtain a new estimated pose, as shown in Fig.~\ref{fig:overall_framwork}. Then ZeroBP uses the new pose to calculate the positional encoding following Eq.~\eqref{eq:coe}. Overall, ZeroBP repeats the above alternate refinement to obtain accurate correspondence between the coarse-level superpoints $\mathcal{P}_{s}^{c}$ and $\mathcal{P}_{o}^{c}$. A similar alternate refinement mechanism is also employed for modeling the correspondence between the fine-level points $\mathcal{P}_{s}^{f}$ and $\mathcal{P}_{o}^{f}$.

\begin{table*}[ht]
  \centering
 \setlength{\tabcolsep}{3.5pt}
\caption{{\bf Pose estimation results on the real-world bin-picking dataset ROBI~\cite{robi}.} 
\note{The AR (\%) of the ADD(-S) metric is reported. In object-specific models, the test object is seen during training, whereas, in zero-shot models, the object is unseen during the training phase. An asterisk (*) indicates asymmetric objects.}}

\begin{adjustbox}{max width=\textwidth}
\begin{tabular}{llcccccccc}
     \toprule
 & & \multicolumn{7}{c}{ROBI Dataset Objects} & \\ 
 \cmidrule(lr){3-9}
     Paradigm & Method & Zigzag* & \makecell[c]{Chrome\\Screw} & Gear & \makecell[c]{Eye\\Bolt} & \makecell[c]{Tube\\Fitting} & \makecell[c]{Din*\\Connector} & \makecell[c]{D-Sub*\\Connector} & {Mean}\\\midrule
     \multirow{5}{*}{\makecell[c]{Object-Specific\\(Seen Object)}} & MIRETR~\cite{miretr}  & 37.6 & 25.7 & 38.0 & 22.2 & 27.2 & 19.9& 11.9 & 24.3\\
     & MP-AAE~\cite{mpaae} & {22.8} & {47.5} & {66.2} & {50.9} & {69.5} & {15.9} & {4.7} & {39.6}\\
     & DC-Net~\cite{dcnet} & {31.0} & {67.7} & {77.6} & {53.5} & {74.8} & {18.7} & {10.6} & {47.4}\\
     & AAE~\cite{aae} & {25.8} & {64.0} & {90.6} & {67.3} & {91.6} & {21.8} & {7.4} & {52.7}\\
     & ST6D~\cite{st6deccv} & {{45.8}} & {{79.9}} & {{97.8}} & {{89.2}} & {{97.5}} & {{24.2}} & {{14.6}} & \textbf{64.1}\\\midrule
     \multirow{3}{*}{\makecell[c]{Zero-Shot\\(Unseen Object)}} & ZeroPose~\cite{chen2023zeropose} & 31.5 & 25.6 & 62.2 & 6.2 & 60.2 & 44.4 & 7.7 & 29.4 \\
     & SAM6D~\cite{sam6d} & 53.4 & 26.8 & 35.3 & 20.4 & 73.1 & 36.7 & 18.9 & 36.7\\ 
     & \textbf{Ours} & 42.8 & 45.6 & 74.4 & 17.5 & 77.8 & 60.1 & 17.5 & \textbf{45.8}\\
     \bottomrule
\end{tabular}
\end{adjustbox}
\label{tab:robi}
\vspace{-1mm}
\end{table*}
\begin{figure*}[!t]
    \centering
    \includegraphics[width=0.9 \textwidth]{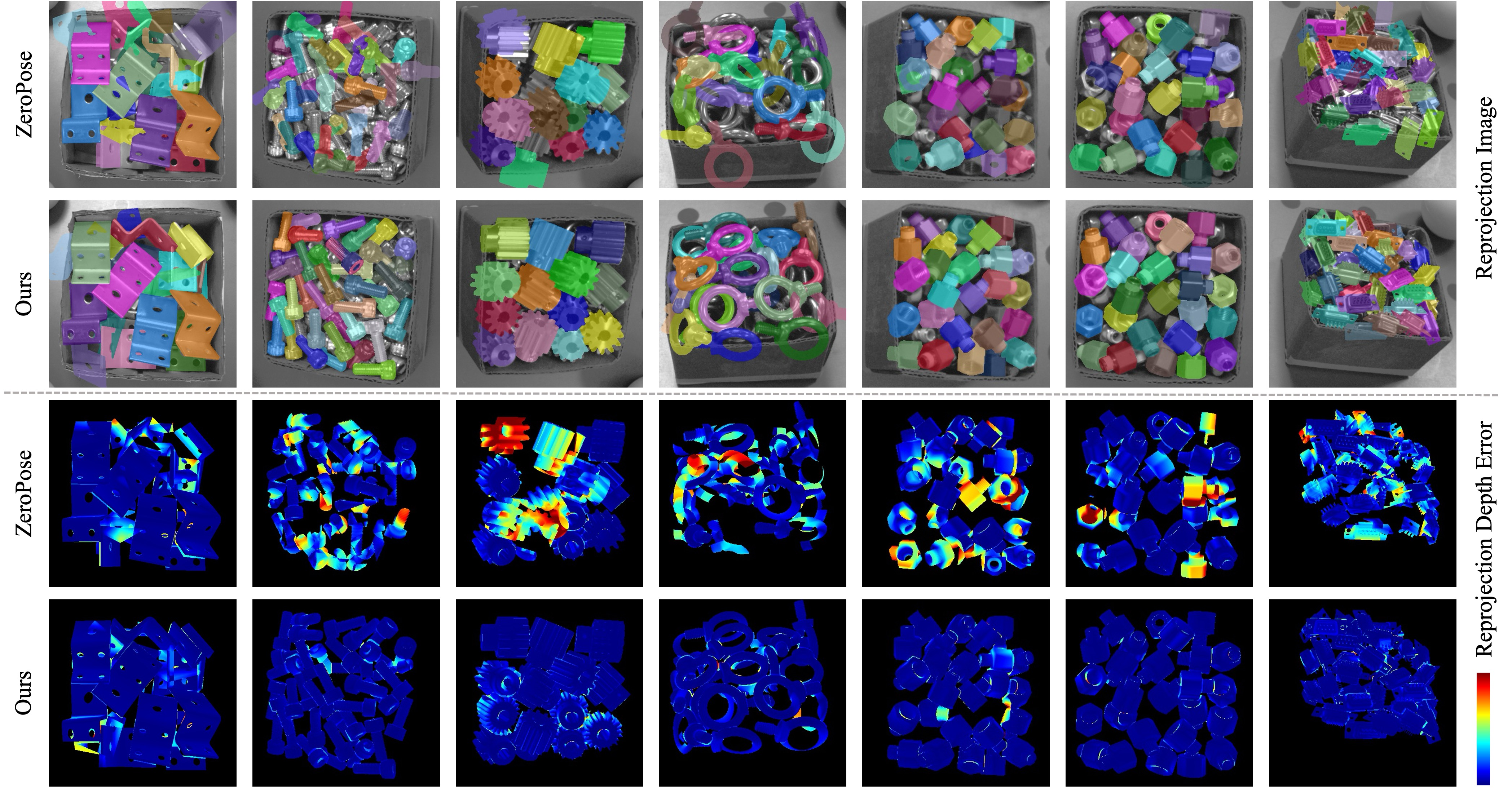}
    \caption{Visualizations of the 6D pose estimation results on the real-world bin-picking dataset ROBI~\cite{robi}.
}
    \label{fig:vis_robi}
    \vspace{-5mm}
\end{figure*}

\subsection{Position-aware cross-attention}
To effectively exploit the global position to improve correspondence modeling, we propose a bidirectional Position-Aware Cross-Attention (PACA) to perform feature interaction between $\mathcal{P}_s^c$ and $\mathcal{P}_o^c$ assisted by the global position information.
As shown in Fig.~\ref{fig:overall_framwork}, PACA leverages the global position information in two manners: 1) it uses an embedding layer to map the positional encodings (\emph{i.e.}, the directional vectors $\hat{\mathcal{V}}_s^c$ and $\hat{\mathcal{V}}_o^c$) into the embedding space and adds the resulting positional embeddings ($\hat{\mathcal{E}}_s^c$ and $\hat{\mathcal{E}}_o^c$) to the superpoint features ($\hat{\mathcal{F}}_s^c$ and $\hat{\mathcal{F}}_o^c$) for attention calculation; and 2) it employs the cosine similarity matrix between the positional encodings $\hat{\mathcal{V}}_s^c$ and $\hat{\mathcal{V}}_o^c$ to reweight the original attention map.

Herein we take the scene-to-CAD cross-attention as an example to formulate the position-aware cross-attention. With the positional embeddings, the query, key, and value are calculated as $Q = (\hat{\mathcal{E}}_s^c + \hat{\mathcal{F}}_s^c)W^q$, $K = (\hat{\mathcal{E}}_o^c + \hat{\mathcal{F}}_o^c)W^k$, $V = \hat{\mathcal{F}}_o^cW^v$, where $W^{q/k/v} \in \mathbb{R}^{d^c \times d^c}$ are projection weights. Then, we use the above-mentioned cosine similarity matrix to reweight the attention map calculated by $K$ and $V$. Thus the reweighted attention map $\mathcal{A}$ can be formulated as:
\begin{align}
    &\mathcal{A} = \sigma(\frac{Q K^T}{\sqrt{d^c}} \cdot \mathcal{A}^c_{\cos}),\\
    &\mathcal{A}_{\cos}^c = (\hat{\mathcal{V}}_s^c (\hat{\mathcal{V}}_o^{c})^{T} + 1) / 2.
\end{align}
Herein $\sigma$ denotes the Softmax function~\cite{lecun1989backpropagation}, $\mathcal{A}_{\cos}^c$ is the cosine similarity matrix normalized to $[0,1]$, and $\cdot$ is element-wise multiplication.
Finally, we multiply the attention matrix $\mathcal{A}$ with the value matrix to obtain the enhanced features $\hat{\mathcal{F}}_s^c$ of the scene superpoints, as follows:
\begin{equation}
    \hat{\mathcal{F}}_s^c = \phi(Q + \mathcal{A}V).
\end{equation}
Herein $\phi$ refers to the feed-forward network~\cite{lecun2015deep}. CAD-to-scene cross-attention can be done in a similar way to obtain the enhanced features $\hat{\mathcal{F}}_o^c$ of the CAD superpoints.

\subsection{Coarse-to-fine position-aware correspondence modeling}
\noindent\textbf{Modeling position-aware correspondence on superpoints.}
To estimate the coarse-level position-aware correspondence in superpoints, we stack the position-aware attention modules \noteb{for} $N$ layers and perform alternate refinement for $N$ steps. 
Given the initial pose $\hat{R}'$ and $\hat{t}'$, and superpoint points $\mathcal{P}_s^c$ and $\mathcal{P}_o^c$, we calculate their position encodings $\hat{\mathcal{V}}_s^c$ and $\hat{\mathcal{V}}_o^c$. Then we combine these positional encodings with their features and perform bidirectional position-aware cross-attention to enhance the superpoint features. After that, we perform matching with the enhanced superpoint features to obtain the correspondence and update the initial pose for the next refinement step. After $N$-step refinement, we obtain a robust coarse-level correspondence.

\vspace{1mm}
\noindent\textbf{Modeling position-aware correspondence on points.}
To estimate the fine-level position-aware correspondence in points, we first find the points in the reception field of superpoints, obtaining the dense points and their fine-level features from the backbone.
Similar to the processing on superpoints, we stack the module for $N$ layers and perform alternate refinement for $N$ steps on dense points. Notably, there are two main differences with the superpoint stage. The initial pose $\hat{R}'$ and $\hat{t}'$ in the first layer is estimated based on the coarse-level correspondence. Following \cite{qin2023geotransformer, chen2023zeropose}, the cross-attention module is removed for efficiency. Herein we update the point feature from the embedding layers. After obtaining the fine-level correspondence, we estimate the final pose $\hat{R}$, $\hat{t}$ by solving Eq.~\eqref{equ:solve_pose}.

\section{Experiments}
\subsection{Experiment dataset} 
\noindent\textbf{Training dataset.} We adopt a large-scaled synthetic dataset GSO~\cite{suresh2023midastouch, labbe2022megapose} to train our model. The synthetic GSO dataset consists of 1000 3D objects and 1 million synthetic RGB-D images with randomly placed objects.

\noindent\textbf{Test dataset.} 
To test the performance in object pose estimation in real-world bin-picking scenes, we adopt the large real-world bin-picking dataset ROBI~\cite{robi} as our test dataset for evaluation. 
This dataset contains seven reflective and texture-less metal objects made from different materials and includes 14 bin-picking test scenes.
These scenes are specifically designed to represent the most challenging scenarios in bin-picking tasks, with up to 38 target instances stacked in a bin.
For each scene, an RGB-D camera captures images from multiple viewpoints, resulting in a total of 1,218 test images with over 20,000 object instances in various poses.

\subsection{Implement details} 
\noindent\textbf{Loss function.} We follow ZeroPose~\cite{chen2023zeropose} and select the overlap-aware circle loss~\cite{qin2023geotransformer}, to optimize the learning PAC on superpoints and negative log-likelihood loss to optimize the learning PAC on points. To enhance correspondence learning in stacked modules, we also optimize the intermediate layer correspondence as an additional loss.

\begin{figure}[t!]
    \centering
    \includegraphics[width=0.49 \textwidth]{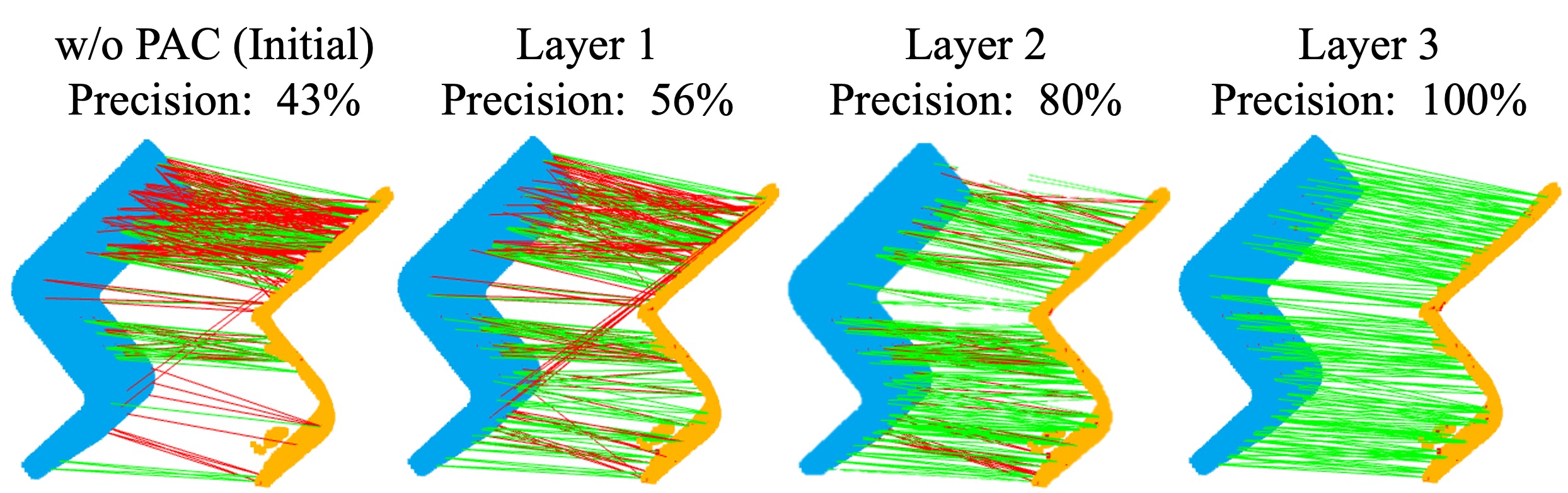}
    \caption{Visualizations of correspondences in layers.}
    \label{fig:vis_layers}
    \vspace{-3mm}
\end{figure}

\vspace{1mm}
\noindent\textbf{Model architecture.}
\note{For fairly compared with the baseline model~\cite{chen2023zeropose}, we keep the same network layers $N=3$ in both superpoints and point correspondence learning.
As early experiments and visualization in Fig.~\ref{fig:vis_layers}, the correspondence can effectively converge with three times alternate refinement.
}

\vspace{1mm}
\noindent\textbf{Correspondences selection.}
\note{We select the superpoint pairs with the top $K=256$ feature similarities as the superpoint correspondences. For point correspondences, we adopt the hypothesis-and-verify approach~\cite{qin2023geotransformer} to estimate the correspondences. This method separately estimates the point correspondences within each superpoint pair to generate pose hypotheses. By transforming all CAD points using these pose hypotheses, point pairs with close distances are considered as the point correspondence. Point correspondences having the largest number in hypotheses are chosen in the final.
}

\subsection{Evaluation metrics} 
We follow \cite{st6deccv, MVBPICRA, dcnet} to select the common average recall (AR) of the ADD(-S) metric~\cite{add} to evaluate the pose estimation performance. This metric evaluates the average point cloud distance between the CAD model transformed by the predicted pose and the ground-truth pose whether smaller than 10\% of the object diameter. 

\begin{table}[t]
    \centering
    \setlength{\tabcolsep}{3.5pt}
    \caption{Comparison of different positional encoding strategies.}
    \begin{adjustbox}{max width=\textwidth}
        \begin{tabular}{rlc}
            \toprule
            & Method & ADD-(S) \\ 
            \midrule
            {\color{teal}\scriptsize 1} & w/o positional encoding & 36.4 \\ 
            {\color{teal}\scriptsize 2} & 3D coordinates &  47.3 \\ 
            {\color{teal}\scriptsize 3} & 3D coordinates with alternate refinement  & 49.3 \\ 
            {\color{teal}\scriptsize 4} & proposed directional vector & 55.7  \\ 
            \bottomrule
        \end{tabular}
    \end{adjustbox}
    \label{tab:aba1_pe_methods}
    \vspace{-2mm}
\end{table}

\begin{table}[t]
    \centering
    \setlength{\tabcolsep}{3.5pt}
    \caption{Effect of our learning PAC in different matching stages.}
    \begin{adjustbox}{max width=\textwidth}
        \begin{tabular}{rccccc}
            \toprule
            & PAC on superpoints & PAC on points & ADD-(S) & Parameters & Runtime \\ 
            \midrule
            {\color{teal}\scriptsize 1} & -- & -- & 36.4 & 9.8M & 0.058\\ 
            {\color{teal}\scriptsize 2} & -- & $\greencheckmark$ & 44.8 & 10.1M & 0.061\\ 
            {\color{teal}\scriptsize 3} & $\greencheckmark$ & --  &  50.2 &  10.1M & 0.071\\ 
            {\color{teal}\scriptsize 4} & $\greencheckmark$ & $\greencheckmark$ & 55.7 & 10.4M & 0.076  \\ 
            \bottomrule
        \end{tabular}
    \end{adjustbox}
    \label{tab:aba_scale}
    \vspace{-2mm}
\end{table}

\subsection{Comparison with state-of-the-art methods}
\noindent\textbf{Quantitative results.}
We compared our model with the baseline ZeroPose~\cite{chen2023zeropose}, and the state-of-the-art zero-shot 6D pose estimation method SAM6D~\cite{sam6d}.
Besides, we also compare with object-specific methods, which need days to generate synthetic data of test objects and retrain the model.
As shown in Table~\ref{tab:robi}, our method outperforms state-of-the-art zero-shot pose estimation methods. Compared with the baseline, it achieves an improvement of 16.4\% in average recall of correct poses.
\noteb{In comparison to object-specific models that have been trained on the test objects, our ZeroBP without seeing these objects during training achieves remarkable performance. While there is still a performance gap, our ZeroBP showcases great potential to narrow the gap.} This generalization capability makes our method a promising alternative to object-specific approaches in real-world bin-picking tasks with faster deployment time and lower deployment cost.

\vspace{1mm}
\noindent\textbf{Qualitative results.}
Fig.~\ref{fig:vis_robi} visualize the 6D pose estimation results of the baseline ZeroPose~\cite{chen2023zeropose} and our method.
For evaluation, we reproject objects into the scene image by the estimated poses and visualize the masks in the top part and the depth error in the bottom part.
To mitigate the impact of potential bias introduced by different segmentation results, we adopt ground truth segmentation results for fair comparison.
As shown in Fig.~\ref{fig:vis_robi}, ZeroPose often leads to significant pose estimation errors in bin-picking objects due to the mismatching in ambiguous regions. Our learning-based PAC method effectively addresses mismatching issues, improving the accuracy of pose estimation.

\subsection{Ablation studies}
We perform ablation studies to analyze our positional encoding strategies and network design. Following~\cite{gcpose}, we conduct the experiments using ground truth masks to eliminate the impact of potential bias in segmentation masks.

\vspace{1mm}
\noindent\textbf{Positional encoding strategies.}
As presented in Tab.~\ref{tab:aba1_pe_methods}, we compare different positional encoding strategies for cross-attention. 
\note{We first select vanilla 3D coordinate positional encoding in cross-attention to distinguish ambiguous regions, which results in a 10.9\% performance gain, demonstrating the effectiveness of positional encoding in heterogeneous feature cross-attention.
The alternate refinement can refine the coordinates and provide an additional 2\% performance improvement.
Benefiting from the natural angle constraint in vector multiplication, the proposed directional vector positional encoding shows the highest AR score 55.7\% and 19.3\% performance gain, demonstrating its effectiveness.}

\vspace{1mm}
\noindent\textbf{Network design.}
As presented in Tab.~\ref{tab:aba_scale}, we compare the pose estimation precision, network parameter size, and runtime at each instance when integrating learning PAC on superpoints and points.
After separately conducting learning PAC on superpoints and points, the AR scores respectively increased by 13.8\% and 8.4\%. When combined together, the AR further improved showing a 19.5\% performance gain.
Notably, the learning-based PAC introduces only a few linear layers for positional embedding, resulting in minimal impact on network speed and parameter size.

\vspace{1mm}
\noindent\textbf{Correspondence alternate refinement in layers.}
\note{Fig.~\ref{fig:vis_layers} shows the visualizations of superpoint correspondences across different layers. 
With the alternate refinement with layers, there is a clear improvement in the precision of the correspondences, starting from 43\% to 100\%. 
The initial correspondences from backbone features show many mismatches within ambiguity regions, indicated by the red lines. With alternate refinement in layers by proposed learning PAC, the number of correct correspondences (green lines) improves significantly, demonstrating its effectiveness.}

\section{Conclusion}
\label{sec:conclusion}
The paper introduces a zero-shot 6D pose estimation method for bin-picking, which learns robust position-aware correspondence between the scene instance and the CAD model to alleviate mismatch in ambiguous regions of textureless bin-picking workpieces.
Thanks to the robust position-aware correspondences, our method offers promising performance in bin-picking scene object pose estimation and strong generalization capability, highlighting its significant potential for more bin-picking applications.


\printbibliography 

\end{document}